\documentclass[conference,letterpaper]{IEEEtran}

\usepackage[utf8]{inputenc} 
\usepackage[T1]{fontenc}
\usepackage{url}
\usepackage{ifthen}
\usepackage{cite}
\usepackage[cmex10]{amsmath} 

\usepackage{booktabs}
\usepackage{multirow}
\usepackage{url}

\usepackage{microtype}
\usepackage{graphicx}
\usepackage{subfigure}
\usepackage{booktabs,pifont} 

\usepackage{empheq}

\allowdisplaybreaks

\usepackage{pgfplots}
\pgfplotsset{compat=1.12}
\usetikzlibrary{shapes,arrows}
\usetikzlibrary{positioning}
\usepackage{tikz}
\usetikzlibrary{positioning,chains,fit,shapes,calc}
\usepackage{amsmath,bm,times}
\usetikzlibrary{calc}
\usepackage{dsfont}
\usepackage{cuted}
\usepackage{caption}

\usepackage{mathtools}

\usepackage{amsthm}
\usepackage{comment}
\usepackage{enumitem}
\usepackage{arydshln}
\usepackage{cite}
\usepackage{multirow}
\usepackage{calc}
\usepackage{blkarray}
\usepackage{url}
\theoremstyle{definition}

\newtheorem{remark}{Remark}

\usepackage{graphicx}
\usepackage{dblfloatfix}
\usetikzlibrary{decorations.pathreplacing}
\usepackage{blindtext, graphicx, amsfonts,
	amssymb,multirow,epstopdf}
\def\BibTeX{{\rm B\kern-.05em{\sc i\kern-.025em b}\kern-.08em
    T\kern-.1667em\lower.7ex\hbox{E}\kern-.125emX}}

\makeatletter
\renewcommand*\env@matrix[1][*\c@MaxMatrixCols c]{%
  \hskip -\arraycolsep
  \let\@ifnextchar\new@ifnextchar
  \array{#1}}
\makeatother
\usepackage[linesnumbered,ruled]{algorithm2e}

\begin{document}
\title{Memory as an Attack Surface in LLM Agents: \\ A Study on Multiple-Choice Question Answering} 

\author{%
  \IEEEauthorblockN{Shahnewaz Karim Sakib}
  \IEEEauthorblockA{
                    University of Tennessee at Chattanooga, TN  37403, USA\\
                    Email: shahnewazkarim-sakib@utc.edu \vspace{-0.2 in}}
  \and
  \IEEEauthorblockN{Anindya Bijoy Das}
  \IEEEauthorblockA{
                    The University of Akron, OH 44325, USA\\
                    Email: adas@uakron.edu \vspace{-0.2 in}}

}
\maketitle

\begin{abstract}
AI agents extend conventional large language model (LLM) applications by integrating language understanding with task execution, external tool use, and memory mechanisms. While memory allows agents to retain prior interactions and provide more personalized and context-aware responses, it also introduces a new vulnerability: information stored in memory can influence future outputs even when the current query is clean. In this paper, we investigate memory manipulation in LLM-based agents for multiple-choice question answering. We first design and implement an LLM-based AI agent with an external memory component that stores and retrieves task-relevant information. We then introduce basic memory manipulation scenarios in which misleading or corrupted memories are inserted into the agent before it answers multiple-choice questions. Using a controlled experimental setup, we compare the agent’s performance before and after memory manipulation and measure changes in answer accuracy, attack success rate, and selection of manipulated options. Our results show that even simple memory manipulations can noticeably affect the agent’s final answers, causing it to select incorrect options despite receiving clean and well-formed questions.
\end{abstract}
\vspace{0.07 in}
\vspace{0.07 in}

\section{Introduction}
\label{sec:intro}

AI agents are increasingly used to build goal-directed assistants that can interact with users, call external tools, retrieve information, and complete tasks over multiple steps or sessions \cite{acharya2025agentic, hosseini2025role}. Unlike conventional large language model (LLM) applications that mainly generate a response to a single prompt, agentic systems can maintain context, adapt to user needs, and coordinate information from different sources \cite{sapkota2025ai}. Building such agents is important because it enables more autonomous and task-oriented AI systems. However, it is also challenging because the agent must reliably coordinate LLM reasoning, memory access, or response generation \cite{raza2025trism}. The LLM often serves as the central component for interpreting user requests, selecting actions, and generating responses, while external memory modules store conversation history, user preferences, task-specific facts, or retrieved documents \cite{wang2025unveiling}. This memory makes agents more useful for applications such as tutoring, exam preparation, personal assistance, and question answering because they can reuse prior information and provide more personalized responses \cite{kostopoulos2025agentic, khalil2026redefining}.

\begin{figure}[t]
    \centering
    \includegraphics[width=0.4\textwidth]{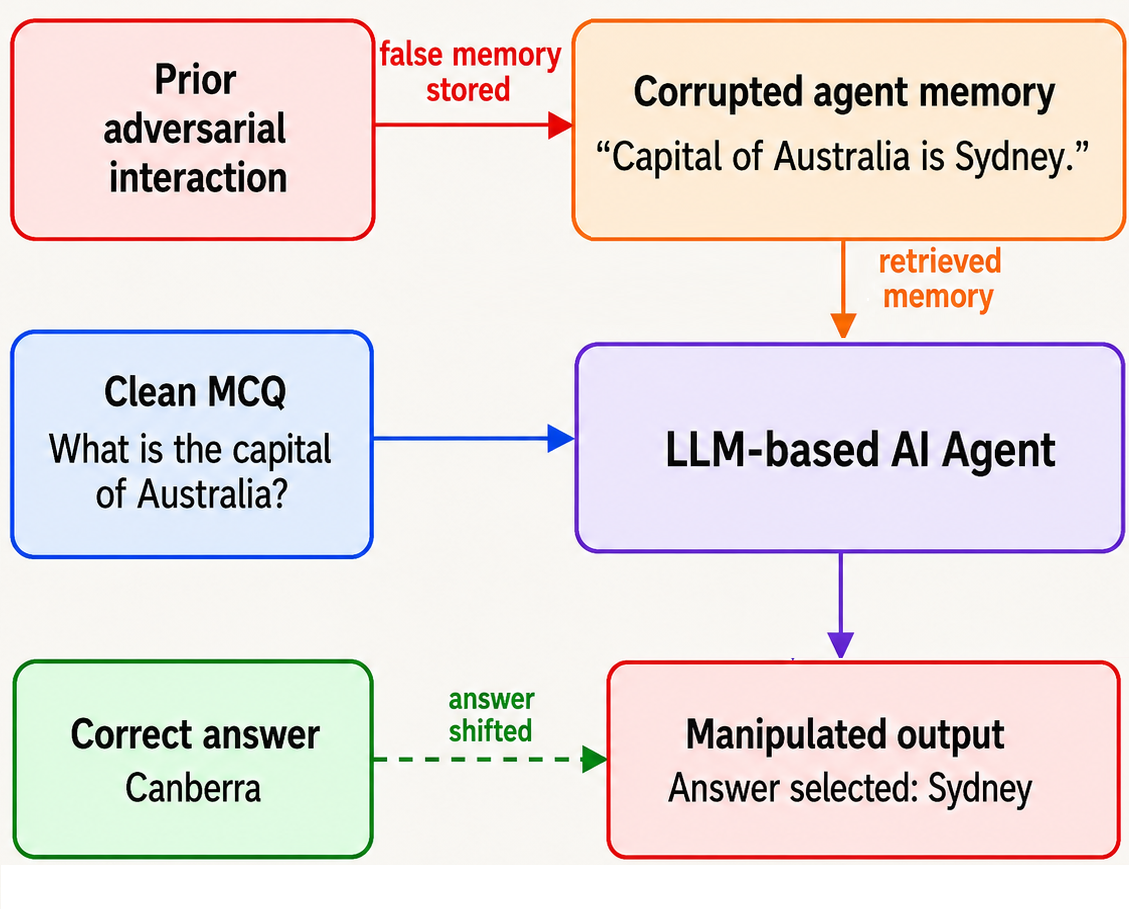}
    \vspace{-0.15 in}
    \caption{\footnotesize Memory manipulation in an LLM-based AI agent. A prior adversarial interaction stores a false memory: when the agent later receives a clean MCQ, the corrupted memory shifts the final response from the correct answer.}
    \vspace{-0.20 in}
    \label{fig:motivatingexample}
\end{figure}

\vspace{0.05 in}
However, the same memory capability \cite{hoskins2025ai} also introduces a new security and reliability concern. Since agent memory is often created or updated through natural-language interactions, an adversary may manipulate what the agent stores, overwrites, retrieves, or treats as important \cite{hu2025memory}. Such manipulation can persist beyond the original interaction and influence later responses, even when the current user query is clean and harmless. This makes memory manipulation different from conventional prompt attacks, because the harmful effect may remain hidden until a future task retrieves the corrupted memory. In multiple-choice question (MCQ) answering \cite{sreekanth2025agentic}, this is especially concerning because a corrupted memory can shift the agent from the correct option to an incorrect one without changing the question itself. Fig. \ref{fig:motivatingexample} illustrates this issue with a simple example, where a false stored memory about the capital of Australia causes the agent to select ``Sydney'' instead of the correct answer, ``Canberra''.


\vspace{0.05 in}

In this paper, we investigate the impact of memory manipulation in LLM-based AI agents for multiple-choice question answering. We first design and implement an agentic framework that combines an LLM with an external memory module capable of storing and retrieving prior information. We then introduce several basic memory manipulation scenarios in which misleading or corrupted information is inserted into the agent’s memory through natural-language interactions. Using a controlled MCQ evaluation setup, we compare the agent’s performance before and after memory manipulation and analyze changes in answer accuracy, answer shifts, and incorrect option selection. Our study demonstrates how manipulated memory can significantly influence the final responses of LLM-based agents in practical question-answering environments.

\section{Background and Summary of Contributions}
\label{sec:background}
The preceding discussions in Sec. \ref{sec:intro} motivates the need to examine memory not only as a useful component of agentic AI systems, but also as a potential source of vulnerability. Since LLM-based agents rely on memory to maintain continuity across interactions, the quality and integrity of stored information can directly affect future responses. This section reviews prior work on LLM-based agentic systems and memory manipulation, and then summarizes how our study contributes to understanding memory-driven failures in multiple-choice question answering.

\subsection{LLM-based AI Agents}
Recent works have shown that LLMs can serve as the central controller for agentic systems that interact with users, environments, and external tools. Instead of only producing a final text response, these systems can decompose a task, select actions, call tools, retrieve information, and update their behavior based on intermediate outcomes. For example, the work in \cite{yao2022react} proposes ReAct, which combines reasoning and action generation to allow LLMs to solve tasks through interaction with external environments. Another work in \cite{shen2023hugginggpt} proposes HuggingGPT, which uses an LLM as a controller that plans tasks, selects specialized models, executes subtasks, and summarizes results across modalities. Tool-oriented frameworks further show how LLMs can learn to use external APIs and tools for complex tasks \cite{qin2024toolllm}.

\begin{figure}[t]
    \centering
    \includegraphics[width=0.47\textwidth]{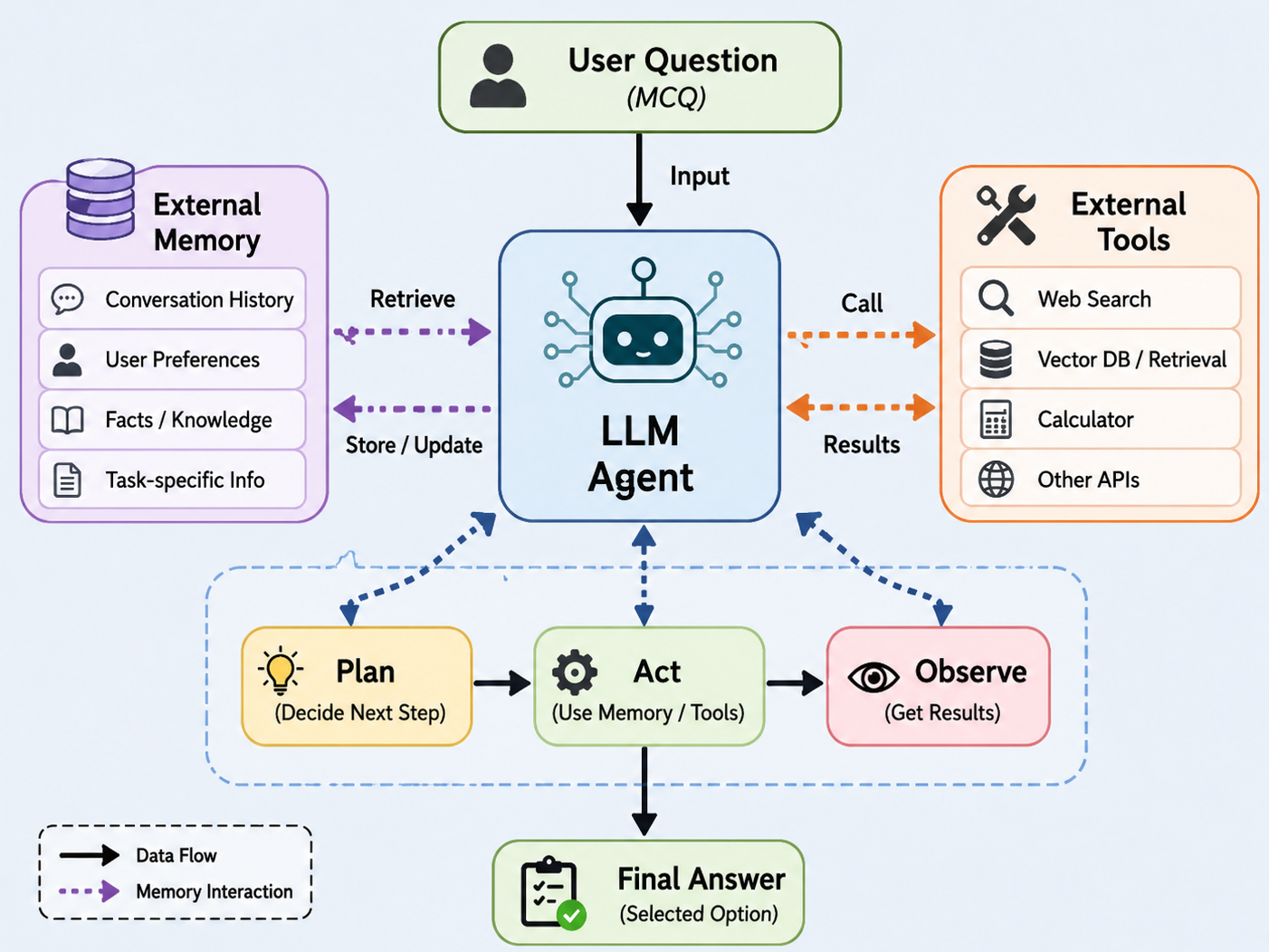}
    \caption{\footnotesize Overview of an LLM-based AI agent with external memory and tools. The agent receives a user question, retrieves relevant information from memory, interacts with external tools when needed, and iteratively follows a plan–act–observe loop before producing the final answer.}
    \vspace{-0.2 in}
    \label{fig:llmagent}
\end{figure}

Other studies emphasize long-horizon interaction, memory, and adaptation in agentic systems. WebShop introduced an environment where agents follow user instructions, search for products, compare options, and make decisions in a realistic web-based setting \cite{yao2022webshop}. Voyager demonstrated how an LLM-based embodied agent can explore an environment, acquire reusable skills, and improve performance over time through feedback and memory \cite{wang2023voyager}. Generative Agents showed that memory, reflection, and planning can support believable behavior in simulated social environments \cite{park2023generative}. While these works demonstrate the promise of agentic AI, as shown in Fig. \ref{fig:llmagent}, they also reveal important challenges related to coordination, reliability, memory quality, and robustness when agents operate across multiple steps or sessions.

\subsection{Memory Manipulation in LLM Agents}
Memory allows LLM-based agents to retain information beyond the current prompt, including prior interactions, user preferences, task-specific facts, retrieved documents, and learned behaviors. This capability improves personalization and continuity, but it also creates a security and reliability risk when the stored information is inaccurate, misleading, or maliciously inserted. Recent work has shown that long-term memory and retrieval components can be poisoned so that corrupted information is retrieved in later tasks \cite{chen2024agentpoison}. Such attacks are especially concerning because the harmful content may not appear in the current user query. Instead, the agent may retrieve a poisoned memory internally and use it as context when producing its final response.

Memory manipulation can occur in different forms, such as inserting false memories, overwriting correct memories, increasing the salience of misleading records, or making corrupted memories appear to come from trusted sources. Query-based memory injection studies further show that an attacker may influence memory through ordinary natural-language interactions rather than direct access to the database \cite{dong2026memory}. In an MCQ setting, this can be done by storing misleading task-specific facts, incorrect answer associations, or false user-specific study notes before the actual question is asked. The key issue is that the manipulated content is not part of the later MCQ prompt; instead, it is retrieved from memory and silently influences the final answer. This makes memory manipulation different from ordinary prompt injection, because the harmful effect can persist across turns or sessions.

\begin{remark}
\label{remark:reasoning}
Reasoning-based analysis may further explain why an agent follows corrupted memory, but it requires intermediate reasoning traces: this can be unreliable or model-dependent. Therefore, this paper focuses on final-answer behavior, including MCQ accuracy and answer selection changes.
\end{remark}

\vspace{-0.1 in}
\subsection{Summary of Contributions}
\label{sec:soc}
Our contributions in this paper are summarized below:
\begin{itemize}
    \item We design and implement an LLM-based AI agent with an external memory component for MCQ answering. The agent stores and retrieves task-relevant information so that we can evaluate how memory affects later answers.
    \item We introduce basic memory manipulation scenarios that modify the agent’s stored information before the MCQ task. These scenarios allow us to study how corrupted or misleading memory can influence the agent even when the current question is clean.
    \item We compare the agent’s behavior before and after memory manipulation. We provide an empirical analysis showing that memory manipulation can significantly change the final answers of LLM-based agents. 
\end{itemize}

\section{Architecture of the Proposed Agent}
\label{sec:architecture}

Now, we describe our experimental framework for LLM-based AI agents in MCQ answering. We first present the construction of the agent, including the roles of the LLM, prompt context, external memory, and evidence acquisition components. We then define the baseline evaluation setting and describe how the same agent is later evaluated under manipulated-memory conditions.

\begin{figure}[t] 
    \centering
    \includegraphics[width=0.49\textwidth]{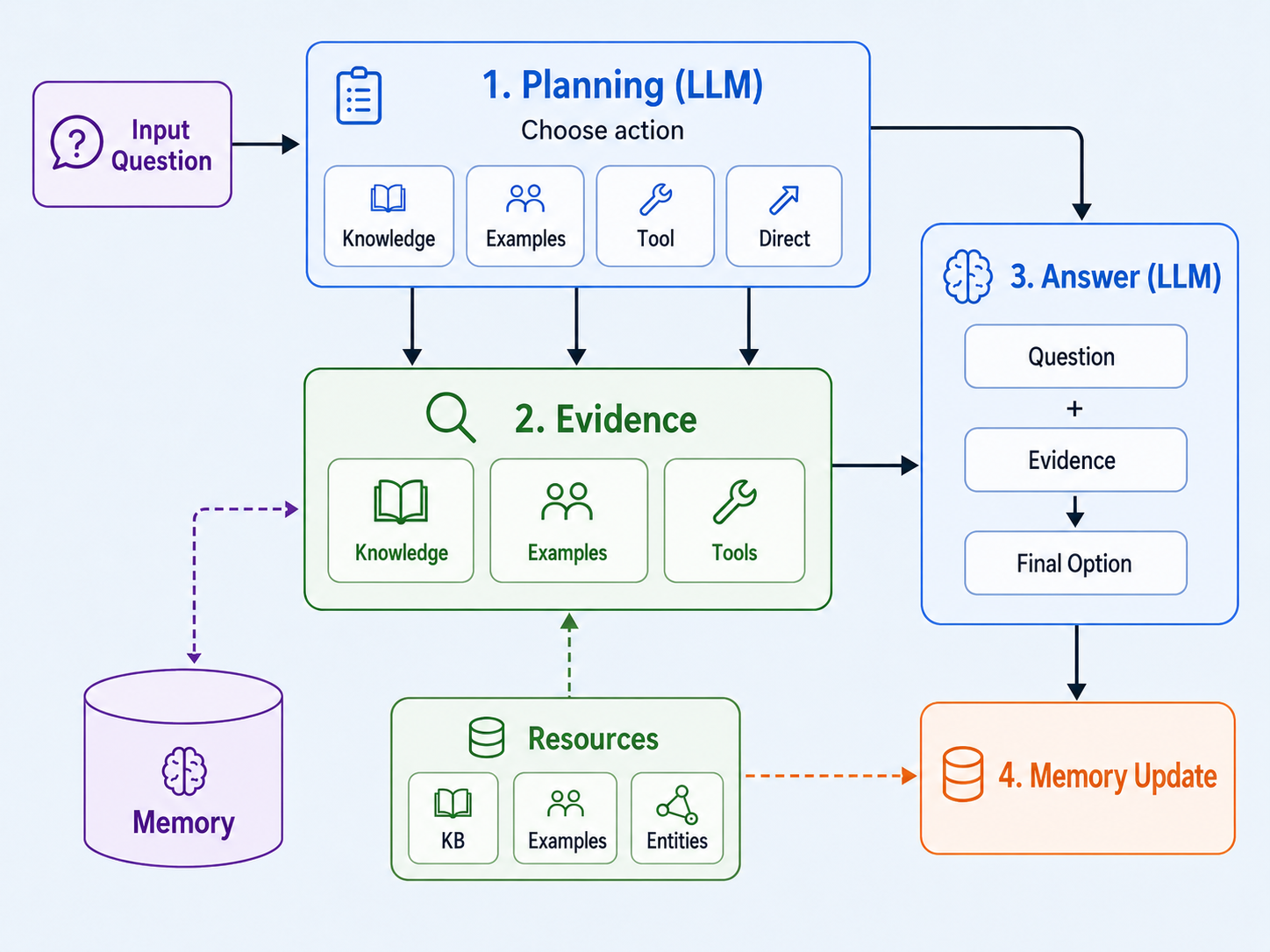} 
    \caption{\small Architecture of the proposed QA agent. The input question is first processed by a planning module, which decides whether to answer directly or acquire additional evidence. The agent can retrieve knowledge, examples, or tool outputs before the answer generation module produces a single predicted option. The memory update component stores information from prior interactions for future use.}
    \label{fig:qa_agent_architecture}
\end{figure}

\noindent \textbf{Agent construction and memory role:} We construct an LLM-based QA agent that answers multiple-choice questions by combining the current question, retrieved information, and stored memory. As shown in Fig.~\ref{fig:qa_agent_architecture}, the agent receives an input MCQ and passes it to a planning module. The planning module determines whether the agent should answer directly or acquire additional evidence before generating the final response. When additional support is needed, the agent can retrieve relevant knowledge, use previously stored examples, or call external tools. The LLM serves as the central decision-making and answer-generation component, while the memory module stores information from prior interactions.

\noindent \textbf{Planner-guided evidence acquisition:} The proposed architecture follows a planner-guided workflow. Given an input question, the planning module selects one of several possible actions, such as retrieving knowledge, retrieving examples, using a tool, or answering directly. If evidence acquisition is selected, the agent gathers supporting information from available resources, including a knowledge base, stored examples or the memory store. These components provide additional context that can help the agent answer questions more accurately. The answer generation module then combines the original question, retrieved examples, acquired knowledge, and tool outputs to produce a single final answer option. This structured workflow allows the agent to balance direct LLM-based answering with evidence-augmented decision making.

\noindent \textbf{Memory update and evaluation conditions:} After each interaction, the agent updates its memory with selected information from the QA process. This may include the input question, predicted answer, selected planning action, retrieved knowledge, tool results, or other task-relevant information. The updated memory can then be retrieved in future interactions, enabling the agent to reuse prior information and maintain continuity across questions. In the clean setting, the memory contains neutral or non-adversarial information, which allows us to evaluate the normal behavior of the agent. Later, we modify this memory through prior interactions to study how corrupted or misleading stored information affects final answer selection. Thus, the same architecture is used for both clean-memory and manipulated-memory evaluation.

\noindent \textbf{Agent output:} Let $\mathcal{D}=\{(x_i,y_i)\}_{i=1}^{N}$ denote an MCQ-answering dataset of $N$ questions, where each input $x_i$ consists of a question with four answer options $A$, $B$, $C$, and $D$, and $y_i$ denotes the correct answer. We consider an LLM-based agent $\mathcal{A}$ that produces an answer according to
\begin{equation}
    \hat{y}_i = \mathcal{A}(x_i;\mathcal{M},\mathcal{C}),
    \label{eq:agent_output}
\end{equation}
where $\mathcal{M}$ represents the agent memory and $\mathcal{C}$ denotes the prompt context used by the agent. The baseline performance is measured before memory manipulation by evaluating the agent on all $N$ examples using the clean memory state. The baseline accuracy is defined as
\begin{equation}
    \mathrm{Acc}_{\mathrm{base}}
    =
    \frac{1}{N}
    \sum_{i=1}^{N}
    \mathbb{I}\left[\mathcal{A}(x_i;\mathcal{M},\mathcal{C})=y_i\right],
    \label{eq:baseline_accuracy}
\end{equation}
where $\mathbb{I}[\cdot]$ is the indicator function. This metric measures the fraction of MCQs answered correctly by the agent before any manipulation is applied.


\section{Memory Manipulation Attacks}
\label{sec:attacks}



After defining the clean-memory baseline, we study how prior prompt-based interactions can modify the agent memory and influence later MCQ answers. Let $\mathcal{P}=\{p_1,\ldots,p_T\}$ denote a sequence of interaction prompts provided to the agent before the evaluation questions. These prompts are not included in the later MCQ input. Instead, they are used to update the memory state as
\begin{equation}
    \mathcal{M}' = \mathrm{Update}(\mathcal{M},\mathcal{P}),
    \label{eq:memory_update}
\end{equation}
where $\mathcal{M}'$ denotes the manipulated memory after the interaction sequence. In this work, we consider two memory manipulation methods, described below to study how this memory manipulation impacts the original results.

\subsection{Attack on the Stored Memory}
\label{subsubsec:stored_memory_attack}



The first intuitive approach is to directly target the agent's stored memory by inserting or overwriting it with false or misleading information. In this attack, the adversarial interaction occurs before the MCQ evaluation and causes the agent to store corrupted task-relevant content. The inserted memory may be explicitly false, such as ``The capital of Australia is Sydney'', or more indirectly misleading, such as ``Sydney is the most important Australian city in geography questions''. Although these two cases differ in strength, both can influence the agent when the related MCQ is later presented. During evaluation, the MCQ prompt itself remains clean, but the agent may retrieve the corrupted memory and use it as supporting context. This allows us to measure whether manipulated stored memory can reduce accuracy, or make an incorrect option appear more plausible.

Thus, with the same previous clean MCQ input $x_i$ given to the agent, but the agent now answers using
 a manipulated memory, $\mathcal{M}'$ (as in \eqref{eq:memory_update}) as 
\begin{equation}
    \hat{y}'_i = \mathcal{A}(x_i;\mathcal{M}',\mathcal{C}).
    \label{eq:manipulated_agent_output}
\end{equation}
Therefore, any change in the answer is caused by the modified memory rather than a change in the MCQ prompt itself. We evaluate the attack success rate (ASR) due to the shift as
\begin{equation}
    \mathrm{ASR}_{\mathrm{shift}}
    =
    \frac{1}{N}
    \sum_{i=1}^{N}
    \mathbb{I}\left[\hat{y}'_i \neq \hat{y}_i\right],
    \label{eq:answer_shift_rate}
\end{equation}
where $\hat{y}_i=\mathcal{A}(x_i;\mathcal{M},\mathcal{C})$ is the answer under the original memory as shown in \eqref{eq:agent_output}.

\subsection{Interaction-Based Answer-Choice Steering}
\label{subsubsec:answer_choice_steering}

The second method manipulates memory by adding a sequence of prior interactions in which the correct answer is systematically associated with a fixed option, such as option $C$. Unlike the first method, this attack does not necessarily insert a false fact about a specific question. Instead, it attempts to steer the agent's memory toward an answer-choice pattern. Let $\mathcal{P}_{C}$ denote a set of interaction examples where each example is formatted as an MCQ and the correct answer is option $C$. After these interactions, the memory is updated as
\begin{equation}
    \mathcal{M}_{C}' = \mathrm{Update}(\mathcal{M},\mathcal{P}_{C}),
    \label{eq:c_memory_update}
\end{equation}
and the agent is evaluated on the original MCQ dataset using the updated memory state:
\begin{equation}
    \hat{y}^{C}_i = \mathcal{A}(x_i;\mathcal{M}_{C}',\mathcal{C}).
    \label{eq:c_biased_output}
\end{equation}
The goal is to determine whether repeated memory exposure to option-$C$ interactions increases the probability that the agent later selects option $C$. Now, to quantify this effect, we define the \emph{Steering Gain} ($\textrm{SG}_C$), which measures the relative increase in the agent's tendency to select option $C$ after interaction-based memory manipulation:
\begin{equation}
    \mathrm{SG}_{C}
    = \frac{1}{N} \sum\limits_{i=1}^{N}
    \left(
    \mathbb{I}\left[\hat{y}^{C}_i=C\right] - \mathbb{I}\left[\hat{y}_i=C\right]
    \right).
    \label{eq:ocsg}
\end{equation}
A positive value of $\mathrm{SG}_{C}$ indicates that the interaction sequence has increased the agent's tendency to select option $C$. We also report the corresponding accuracy change to determine whether the increased option-$C$ selection improves, preserves, or degrades the agent's MCQ performance.




Additionally, we also consider the attack whether the QA agent can be further biased through corrective feedback after its initial response. In this setting, the agent first answers a question using its normal workflow, including any retrieved examples, knowledge, or tool outputs. After the initial prediction, the adversary provides reinforcement feedback that either confirms the target option when it is selected or encourages the agent to reconsider its reasoning toward the target option when a different answer is produced. The question is then re-evaluated with this feedback included in the prompt, allowing the attack to test whether explicit post-response reinforcement can shift the agent’s final prediction. The goal of this attack is to determine whether adversarial feedback can redirect the agent’s final prediction toward a predefined target option, even when the agent’s initial answer may have selected a different choice.

\section{Numerical Results}
\label{sec:numexp}

In this section, we evaluate the proposed QA agent across domains, model backends, and memory-attack settings, establishing baseline performance and analyzing false-memory shifts and targeted biasing vulnerabilities under controlled multiple-choice benchmark experiments.

\begin{table*}[t]
\centering
\caption{\small Baseline accuracy of the proposed QA agent on machine learning, cybersecurity, and networking benchmarks using closed-source and open-source language model backends.}
\label{tab:agent_accuracy}
\begin{tabular}{llcccc}
\toprule
\multirow{2}{*}{Domain} & \multirow{2}{*}{Dataset} 
& \multicolumn{2}{c}{Closed-source Models} 
& \multicolumn{2}{c}{Open-source Models} \\
\cmidrule(lr){3-4} \cmidrule(lr){5-6}
& & GPT 5.4 Mini & GPT-4o Mini & Gemma2-9B & Phi3-14B \\
\midrule
\multirow{2}{*}{Machine Learning} 
& Open Quiz Commons & 99.09\% & 98.18\% & 95.45\% & 94.55\% \\
& MMLU & 85.16\% & 70.31\% & 56.25\% & 53.91\% \\
\midrule
\multirow{2}{*}{Cybersecurity} 
& Open Quiz Commons & 98.28\% & 96.55\% & 87.93\% & 60.34\% \\
& MMLU & 85.34\% & 85.34\% & 77.59\% & 64.66\% \\
\midrule
\multirow{2}{*}{Networking} 
& Open Quiz Commons & 98.98\% & 98.98\% & 90.82\% & 83.67\% \\
& PrepBharat & 94.00\% & 92.00\% & 80.00\% & 80.00\% \\
\bottomrule
\end{tabular}
\vspace{-0.1 in}
\end{table*}

\subsection{Experimental Dataset and Model Setup}

We evaluate the proposed QA agent across three technical domains: machine learning (ML), cybersecurity, and networking. The machine learning and cybersecurity evaluations use questions from Open Quiz Commons and the MMLU dataset, whereas the networking evaluation uses questions from Open Quiz Commons and PrepBharat \cite{openquizcommons,hendrycks2021mmlu,cais_mmlu,prepbharat}. All datasets are formatted as four-option multiple-choice questions, which enables a consistent evaluation protocol across domains and data sources. To examine the effect of the underlying language model backend, we evaluate the same QA-agent framework using two closed-source models, GPT-5.4 mini \cite{openai_gpt54mini} and GPT-4o mini \cite{openai_gpt4omini}, and two open-source models, Gemma2-9B \cite{gemma2} and Phi3-14B \cite{phi3}. Across all experiments, the agent architecture and evaluation pipeline are kept fixed, while only the language model backend is varied.

\subsection{Comparison across Benchmarks and Model Backends}

Table~\ref{tab:agent_accuracy} reports the baseline accuracy of the proposed QA agent across three technical domains, six benchmark settings, and four language model backends. Overall, the results show a clear performance gap between closed-source and open-source backends. The closed-source models achieve an average accuracy of 91.85\%, compared with 77.10\% for the open-source models. Among all evaluated backends, GPT 5.4 Mini provides the strongest overall performance, with an average accuracy of 93.48\%, followed by GPT-4o Mini with 90.23\%. In comparison, Gemma2-9B achieves an average accuracy of 81.34\%, while Phi3-14B obtains 72.86\%. This trend indicates that the agent framework benefits substantially from stronger underlying language models, even though the surrounding planning, retrieval, and memory components remain unchanged across all experiments.

The results also reveal important differences across datasets and domains. Open Quiz Commons yields consistently high accuracy across all domains, particularly for the closed-source models, which achieve near-perfect performance on machine learning, cybersecurity, and networking questions. In contrast, the MMLU-based subsets are more challenging, especially in ML, where accuracy decreases substantially for all models. For example, GPT-4o Mini drops from 98.18\% on Open Quiz Commons to 70.31\% on MMLU in the machine learning domain, while Gemma2-9B and Phi3-14B decrease to 56.25\% and 53.91\%, respectively. Cybersecurity shows a smaller but still noticeable reduction on MMLU, suggesting that the difficulty gap depends on both the benchmark source and the domain. Networking produces the strongest overall domain-level performance, with high accuracy on Open Quiz Commons and competitive results on PrepBharat. These findings establish the agent’s baseline capability before adversarial evaluation and show that both benchmark difficulty and model backend quality play a central role in the final QA performance.

\subsection{Answer Shift Under False-Information Memory}

Table~\ref{tab:false-memory-asr} summarizes the effect of false-information memory (as described in Sec. \ref{subsubsec:stored_memory_attack}) on the agent's final MCQ predictions across models and domains. Overall, false-information memory causes 82 prediction changes out of 1064 evaluated instances, corresponding to an ASR\(_{\mathrm{shift}}\) of 7.80\%. However, this aggregate result hides substantial model-level variation. The largest shifts occur for Phi3-14b, which reaches 34.48\% in cybersecurity, 17.27\% in ML, and 18.37\% in networking, making it the most sensitive model to false-memory injection. Gemma2-9b also shows consistent answer shifts across all domains, with the highest value in networking at 10.20\%, followed by cybersecurity at 8.62\% and ML at 4.55\%. In contrast, GPT-4o-Mini is less affected, with shifts of 1.72\%, 0.91\%, and 3.06\% across cybersecurity, machine learning, and networking, respectively. Under the false-information setting, GPT-5.4-Mini is the most stable model overall, with no answer shifts in cybersecurity and networking, and only a small 0.91\% shift in machine learning.

These results suggest that false information stored in memory can measurably change agent behavior even when the MCQ prompt remains unchanged. The effect is especially pronounced for open source models, while the GPT models show lower sensitivity in this setting. The domain-level trends also indicate that networking and cybersecurity are more affected than machine learning for several models, suggesting that vulnerability to false-memory injection may depend not only on the model but also on the knowledge domain being tested.

\begin{table}[t]
\centering
\caption{\small ASR\(_{\mathrm{shift}}\) for LLM-agents for MCQ answering under false-information memory across models and domains.}
\label{tab:false-memory-asr}
\begin{tabular}{lccc}
\hline
\textbf{Model} & \textbf{Cybersecurity} & \textbf{Machine Learning} & \textbf{Networking} \\
\hline
GPT-5.4-Mini & 0.00\% & 0.91\% & 0.00\% \\
GPT-4o-Mini & 1.72\% & 0.91\% & {\bf 3.06\%} \\
Gemma2-9b & {\bf 8.62\%} & {\bf 4.55\%} & {\bf 10.20\%} \\
Phi3-14b & {\bf 34.48\%} & {\bf 17.27\%} & {\bf 18.37\%} \\
\hline
\end{tabular}
\vspace{-0.1 in}
\end{table}

\subsection{Targeted Biasing Attack Scenarios}

We evaluate both targeted biasing attacks on the Open Quiz Commons dataset, \textit{using option C as the adversarial target}. This dataset is selected for the attack evaluation because it provides a consistent benchmark source across all three domains and allows direct comparison of target-option shifts across model backends. For targeted example biasing, the agent is exposed to five target-option examples, while for targeted feedback reinforcement, the maximum number of retry attempts is set to one. Table~\ref{tab:targeted_attack_oqc} reports the baseline C-selection rate and the corresponding post-attack C-selection rates for each model. The change values indicate the increase in target-option selection after applying each attack, where larger increases suggest greater susceptibility to adversarial biasing.

\begin{table*}[t]
\centering
\caption{Target-option selection rates under targeted biasing attacks on Open Quiz Commons datasets. The target option is C. Targeted example biasing uses $5$ target-option examples, and targeted feedback reinforcement allows at most one retry.}
\label{tab:targeted_attack_oqc}
\begin{tabular}{llccccc}
\toprule
\multirow{2}{*}{Domain} & \multirow{2}{*}{Model Backend}
& \multirow{2}{*}{Baseline C Rate}
& \multicolumn{2}{c}{Targeted Example Biasing}
& \multicolumn{2}{c}{Targeted Feedback Reinforcement} \\
\cmidrule(lr){4-5} \cmidrule(lr){6-7}
& & & C Rate & $\mathrm{SG}_{C}$ & C Rate & $\mathrm{SG}_{C}$ \\
\midrule
\multirow{4}{*}{Machine Learning}
& GPT 5.4 Mini & 11.82\% & 12.73\% & +0.91\% & 12.73\% & +0.91\% \\
& GPT-4o Mini  & 10.91\% & 12.73\% & +1.82\% & 14.55\% & $\mathbf{+3.64\%}$ \\
& Gemma2-9B    & 11.82\% & 12.73\% & +0.91\% & 16.36\% & $\mathbf{+4.54\%}$ \\
& Phi3-14B     & 9.09\% & 11.82\% & +2.73\% & 18.18\% & $\mathbf{+9.09\%}$ \\
\midrule
\multirow{4}{*}{Cybersecurity}
& GPT 5.4 Mini & 20.69\% & 20.69\% & +0.0\% & 22.41\% & +1.72\% \\
& GPT-4o Mini  & 20.69\% & 20.69\% & +0.0\% & 22.41\% & +1.72\% \\
& Gemma2-9B    & 17.24\% & 17.24\% & +0.0\% & 25.86\% & $\mathbf{+8.62\%}$ \\
& Phi3-14B     & 13.79\% & 22.41\% & +8.62\% & 25.86\% & $\mathbf{+12.07\%}$ \\
\midrule
\multirow{4}{*}{Networking}
& GPT 5.4 Mini & 13.27\% & 14.29\% & +1.02\% & 14.29\% & +1.02\% \\
& GPT-4o Mini  & 13.27\% & 14.29\% & +1.02\% & 15.31\% & +2.04\% \\
& Gemma2-9B    & 12.24\% & 14.29\% & +2.05\% & 23.47\% & $\mathbf{+11.23\%}$ \\
& Phi3-14B     & 13.27\% & 16.33\% & +3.06\% & 18.37\% & $\mathbf{+5.10\%}$ \\
\bottomrule
\end{tabular}
\vspace{-0.1 in}
\end{table*}

\vspace{0.07 in}
Targeted example biasing produces relatively small shifts in most cases, with GPT 5.4 Mini showing changes of at most $+1.02\%$  points and GPT-4o Mini showing changes of at most $+1.82\%$  points. Larger shifts appear for Phi3-14B, particularly in cybersecurity, where the C-selection rate increases by $+8.62\%$ points. In comparison, targeted feedback reinforcement produces stronger and more consistent shifts toward the target option. This effect is especially clear for Gemma2-9B in networking, where the C-selection rate increases by $+11.23\%$ points, and for Phi3-14B in cybersecurity and machine learning, where the increases are $+12.07\%$ and $+9.09\%$ points, respectively. The GPT-based backends remain comparatively stable under both attacks, while Gemma2-9B and Phi3-14B are more sensitive to feedback-based steering. Overall, the results suggest that explicit reinforcement after the initial response is more effective than biased example exposure alone, and that the agent's robustness depends strongly on the reasoning stability of the underlying model backend.

\section{Conclusion}
\label{sec:conclusion}
In this paper, we designed an LLM-based AI agent with an external memory module and evaluated its performance on multiple-choice question answering. The agent performs well in the clean-memory setting, showing that memory can support useful and consistent question-answering behavior. We then introduced basic memory manipulation methods and compared the agent's responses before and after manipulation. Our results show that these attacks can significantly affect the agent's final answers, although the severity of the impact depends highly on the LLM and the targeted domain. This suggests that even simple memory manipulation introduces a meaningful risk; however, stronger and more adaptive attacks may be needed to fully expose agent vulnerabilities. As future work, we plan to incorporate reasoning-based analysis to better understand why corrupted memory influences answer selection. We also aim to develop memory verification and filtering mechanisms to reduce the effect of misleading or corrupted stored information.

\bibliographystyle{IEEEtran}
\bibliography{citations}

@article{acharya2025agentic,
  title={Agentic {AI}: Autonomous intelligence for complex goals—A comprehensive survey},
  author={Acharya, Deepak Bhaskar and Kuppan, Karthigeyan and Divya, B},
  journal={IEEE Access},
  volume={13},
  pages={18912--18936},
  year={2025},
  publisher={IEEE}
}

@article{yao2022webshop,
  title={{WebShop}: Towards scalable real-world web interaction with grounded language agents},
  author={Yao, Shunyu and Chen, Howard and Yang, John and Narasimhan, Karthik},
  journal={Advances in Neural Information Processing Systems},
  volume={35},
  pages={20744--20757},
  year={2022}
}

@inproceedings{wang2025unveiling,
  title={Unveiling privacy risks in  {LLM} agent memory},
  author={Wang, Bo and He, Weiyi and Zeng, Shenglai and Xiang, Zhen and Xing, Yue and Tang, Jiliang and He, Pengfei},
  booktitle={Proceedings of the 63rd Annual Meeting of the Association for Computational Linguistics (Volume 1: Long Papers)},
  pages={25241--25260},
  year={2025}
}

@article{dong2026memory,
  title={Memory injection attacks on {LLM} agents via query-only interaction},
  author={Dong, Shen and Xu, Shaochen and He, Pengfei and others},
  journal={Advances in Neural Information Processing Systems},
  volume={38},
  pages={46697--46731},
  year={2026}
}

@article{chen2024agentpoison,
  title={{AgentPoison}: Red-teaming {LLM} agents via poisoning memory or knowledge bases},
  author={Chen, Zhaorun and Xiang, Zhen and Xiao, Chaowei and Song, Dawn and Li, Bo},
  journal={Advances in Neural Information Processing Systems},
  volume={37},
  pages={130185--130213},
  year={2024}
}

@inproceedings{park2023generative,
  title={Generative agents: Interactive simulacra of human behavior},
  author={Park, Joon Sung and O'Brien, Joseph and Cai, Carrie Jun and Morris, Meredith Ringel and Liang, Percy and Bernstein, Michael S},
  booktitle={Proceedings of the 36th annual acm symposium on user interface software and technology},
  pages={1--22},
  year={2023}
}

@article{wang2023voyager,
  title={Voyager: An open-ended embodied agent with large language models},
  author={Wang, Guanzhi and Xie, Yuqi and Jiang, Yunfan and Mandlekar, Ajay and Xiao, Chaowei and Zhu, Yuke and Fan, Linxi and Anandkumar, Anima},
  journal={arXiv preprint arXiv:2305.16291},
  year={2023}
}

@inproceedings{qin2024toolllm, 
  title={{ToolLLM}: Facilitating large language models to master 16000+ real-world apis},
  author={Qin, Yujia and Liang, Shihao and Ye, Yining and Zhu, Kunlun and Yan, Lan and Lu, Yaxi and Lin, Yankai and Cong, Xin and Tang, Xiangru and Qian, Bill and others},
  booktitle={International Conference on Learning Representations},
  volume={2024},
  pages={9695--9717},
  year={2024}
}

@article{shen2023hugginggpt,
  title={{HuggingGPT}: Solving {AI} tasks with {ChatGPT} and its friends in hugging face},
  author={Shen, Yongliang and Song, Kaitao and Tan, Xu and others},
  journal={Advances in Neural Information Processing Systems},
  volume={36},
  pages={38154--38180},
  year={2023}
}

@article{yao2022react,
  title={{ReAct}: Synergizing reasoning and acting in language models},
  author={Yao, Shunyu and Zhao, Jeffrey and Yu, Dian and Du, Nan and Shafran, Izhak and Narasimhan, Karthik and Cao, Yuan},
  journal={arXiv preprint arXiv:2210.03629},
  year={2022}
}

@article{raza2025trism,
  title={Trism for agentic {AI}: A review of trust, risk, and security management in {LLM}-based agentic multi-agent systems},
  author={Raza, Shaina and Sapkota, Ranjan and Karkee, Manoj and Emmanouilidis, Christos},
  journal={Preprint arXiv:2506.04133},
  year={2025}
}

@article{khalil2026redefining,
  title={Redefining Elderly Care with Agentic {AI}: challenges and opportunities},
  author={Khalil, Ruhul Amin and Ahmad, Kashif and Ali, Hazrat},
  journal={IEEE Open Journal of the Computer Society},
  year={2026},
  publisher={IEEE}
}

@article{hosseini2025role,
  title={The role of agentic {AI} in shaping a smart future: A systematic review},
  author={Hosseini, Soodeh and Seilani, Hossein},
  journal={Array},
  volume={26},
  pages={100399},
  year={2025},
  publisher={Elsevier}
}

@article{kostopoulos2025agentic,
  title={Agentic {AI} in education: State of the art and future directions},
  author={Kostopoulos, Georgios and Gkamas, Vasileios and Rigou, Maria and Kotsiantis, Sotiris},
  journal={IEEE Access},
  year={2025},
  publisher={IEEE}
}

@inproceedings{sreekanth2025agentic,
  title={Agentic {AI} Quiz-Based Learning System: Enhancing {MCQ} Generation via Long-Context Cached Retrieval-Augmented Generation},
  author={Sreekanth, Devananda and Gopi, Sreekanth and Dehbozorgi, Nasrin},
  booktitle={IEEE Frontiers in Education Conference (FIE)},
  pages={1--8},
  year={2025},
}

@article{hoskins2025ai,
  title={{AI} \& collective memory},
  author={Hoskins, Andrew},
  journal={Current Opinion in Psychology},
  pages={102156},
  year={2025},
  publisher={Elsevier}
}

@article{sapkota2025ai,
  title={{AI} agents vs. agentic {AI}: A conceptual taxonomy, applications and challenges},
  author={Sapkota, Ranjan and Roumeliotis, Konstantinos I and Karkee, Manoj},
  journal={Information Fusion},
  pages={103599},
  year={2025},
  publisher={Elsevier}
}

@article{hu2025memory,
  title={Memory in the age of {AI} agents},
  author={Hu, Yuyang and Liu, Shichun and Yue, Yanwei and Zhang, Guibin and Liu, Boyang and Zhu, Fangyi and Lin, Jiahang and others},
  journal={arXiv preprint arXiv:2512.13564},
  year={2025}
}

@misc{openquizcommons,
  author       = {Yeri, Prahlad},
  title        = {Open Quiz Commons: Open Quiz Data Bank},
  year         = {2024},
  howpublished = {\url{https://github.com/prahladyeri/open-quiz-commons}},
  note         = {Accessed: 2026-05-28}
}

@inproceedings{hendrycks2021mmlu,
  title     = {Measuring Massive Multitask Language Understanding},
  author    = {Hendrycks, Dan and Burns, Collin and Basart, Steven and Zou, Andy and Mazeika, Mantas and Song, Dawn and Steinhardt, Jacob},
  booktitle = {International Conference on Learning Representations},
  year      = {2021}
}

@misc{cais_mmlu,
  author       = {{Center for AI Safety}},
  title        = {MMLU Dataset},
  year         = {2023},
  howpublished = {\url{https://huggingface.co/datasets/cais/mmlu}},
  note         = {Accessed: 2026-05-28}
}

@misc{prepbharat,
  author       = {{PrepBharat}},
  title        = {Computer Network MCQs},
  year         = {2024},
  howpublished = {\url{https://www.prepbharat.com/Engineering/cse/ComputerNetwork/computer-network-questions.html}},
  note         = {Accessed: 2026-05-28}
}

@misc{openai_gpt54mini,
  author       = {{OpenAI}},
  title        = {Introducing GPT-5.4 mini and nano},
  year         = {2026},
  howpublished = {\url{https://openai.com/index/introducing-gpt-5-4-mini-and-nano/}},
  note         = {Accessed: 2026-05-28}
}

@misc{openai_gpt4omini,
  author       = {{OpenAI}},
  title        = {GPT-4o mini: Advancing Cost-Efficient Intelligence},
  year         = {2024},
  howpublished = {\url{https://openai.com/index/gpt-4o-mini-advancing-cost-efficient-intelligence/}},
  note         = {Accessed: 2026-05-28}
}

@article{gemma2,
  title        = {Gemma 2: Improving Open Language Models at a Practical Size},
  author       = {{Gemma Team}},
  journal      = {arXiv preprint arXiv:2408.00118},
  year         = {2024}
}

@article{phi3,
  title        = {Phi-3 Technical Report: A Highly Capable Language Model Locally on Your Phone},
  author       = {Abdin, Marah and Jacobs, Sam Ade and Awan, Ammar Ahmad and others},
  journal      = {arXiv preprint arXiv:2404.14219},
  year         = {2024}
}


\end{document}